\def\year{2022}\relax
\documentclass[letterpaper]{article} % DO NOT CHANGE THIS
\usepackage{aaai22}  % DO NOT CHANGE THIS
\usepackage{times}  % DO NOT CHANGE THIS
\usepackage{helvet}  % DO NOT CHANGE THIS
\usepackage{courier}  % DO NOT CHANGE THIS
\usepackage[hyphens]{url}  % DO NOT CHANGE THIS
\usepackage{graphicx} % DO NOT CHANGE THIS
\usepackage{natbib}  % DO NOT CHANGE THIS AND DO NOT ADD ANY OPTIONS TO IT
\usepackage{caption} % DO NOT CHANGE THIS AND DO NOT ADD ANY OPTIONS TO IT
\DeclareCaptionStyle{ruled}{labelfont=normalfont,labelsep=colon,strut=off} % DO NOT CHANGE THIS
\usepackage{algorithm}
\usepackage{algorithmic}
\usepackage{subfiles}
\usepackage{color}
\usepackage{amsmath}
\usepackage{amssymb}
\usepackage{multirow}
\usepackage{comment}
\usepackage{booktabs} % for professional tables
\usepackage{tabularx}
\usepackage{array}
\newcolumntype{L}[1]{>{\raggedright\let\newline\\\arraybackslash\hspace{0pt}}m{#1}}
\newcolumntype{C}[1]{>{\centering\let\newline\\\arraybackslash\hspace{0pt}}m{#1}}
\newcolumntype{R}[1]{>{\raggedleft\let\newline\\\arraybackslash\hspace{0pt}}m{#1}}

\usepackage{newfloat}
\usepackage{listings}
\floatstyle{ruled}
\newfloat{listing}{tb}{lst}{}
\floatname{listing}{Listing}
\newcommand{\todo}[2]{{\color{blue}@#1:#2}}

\title{Novelty Controlled Paraphrase Generation with Retrieval Augmented Conditional Prompt Tuning}
\author {
    Jishnu Ray Chowdhury,\textsuperscript{\rm 1}\footnote{The work was done during an internship with Bloomberg. Correspondence to: Jishnu Ray Chowdhury, Yong Zhuang, Shuyi Wang.}
    Yong Zhuang, \textsuperscript{\rm 2}
    Shuyi Wang \textsuperscript{\rm 2} \\
}
\affiliations {
    \textsuperscript{\rm 1} University of Illinois, at Chicago\\
    \textsuperscript{\rm 2} Bloomberg\\
    jraych2@uic.edu, yzhuang52@bloomberg.net, swang1072@bloomberg.net
}

\begin{document}

\maketitle

\begin{abstract}
Paraphrase generation is a fundamental and long-standing task in natural language processing. In this paper, we concentrate on two contributions to the task: (1) we propose Retrieval Augmented Prompt Tuning (RAPT) as a parameter-efficient method to adapt large pre-trained language models for paraphrase generation; (2) we propose
Novelty Conditioned RAPT (NC-RAPT) as a simple model-agnostic method of using specialized prompt tokens for controlled paraphrase generation with varying levels of lexical novelty. By conducting extensive experiments on four datasets, we demonstrate the effectiveness of the proposed approaches for retaining the semantic content of the original text while inducing lexical novelty in the generation.
\end{abstract}

\section{Introduction}
\label{sec-introduction}
The task of paraphrase generation aims at rephrasing a given text while retaining its meaning. The task has several applications including text simplification, semantic parsing, query re-writing, and data augmentation. 

%In this work, first, we focus on adaptation of pre-trained language models for paraphrase generation. Second, we focus on improving the novelty of the generated paraphrases. 

Recently, the use of pre-trained Transformer-based language models has become nearly ubiquitous for different natural language processing (NLP) tasks \cite{devlin2019bert, radford2019language}  including paraphrase generation \cite{witteveen-andrews-2019-paraphrasing, west2021reflective}. While the standard method of utilizing a pre-trained language model for NLP tasks is to fine-tune all the parameters in the model, it is not the most parameter-efficient. For example, MegatronLM \cite{shoeybi2019megatron} has more than 11 billion parameters to fine-tune.
Various methods have emerged to utilize large-scale pre-trained language models in a more parameter-efficient manner. Such methods include variants of adapter tuning \cite{houlsby2019parameter}, prompt tuning \cite{shin2020autoprompt, li2021prefix, lester2021power}, and Low Rank Adaptation (LoRA) \cite{hu2021lora}. Particularly, prompt tuning and LoRA can cut down the number of trainable parameters by a factor of thousands. %Such methods allow memory-efficient training and 
%They require storage of only a few task-specific parameters for each task. 
These approaches are particularly important today as pre-trained models continuously become larger than before. In the direction of parameter-efficient methods, we propose Retrieval Augmented Prompt Tuning (RAPT) as a method of augmenting prompt tuning with kNN-based retrieved examples to further enhance the quality of paraphrases.  

Besides adaptation of pre-trained language models, we also explicitly focus on the novelty of generated paraphrases because trivial paraphrases with minimal changes may not be as helpful for applications such as data augmentation.
We consider the novelty of a paraphrase to be associated with the edit distance of the paraphrase from the input. Roughly, the more edits (word deletion, word insertion, substitution, etc.) there are in a paraphrase, the more ``novel" we consider it to be. %Addition of new words, deletion of unnecessary words, substitution, or swapping, all count as different ways of editing an input text for paraphrasing. 
Paraphrases can be generated with different levels of novelty. For example, simply changing one word would still count as paraphrasing. Usually, however, the level of lexical novelty of a generated paraphrase is left upon the model to decide. In this work, we instead propose a simple model-agnostic method to put more control on the users themselves to decide the level of novelty of the generation. To this end, we propose Novelty Conditioned RAPT (NC-RAPT) that uses specialized prompts for different levels of novelty for novelty-controlled paraphrase generation. Overall, we make the following contributions:
\begin{enumerate}
    \item We propose Retrieval Augmented Prompt Tuning (RAPT) as a parameter-efficient method for paraphrase generation.
    \item We propose Novelty Conditioned RAPT (NC-RAPT) that uses specialized prompts to generate paraphrases with different levels of novelty. 
\end{enumerate}

\section{Task Definition}
\label{sec-task-definition}
We define paraphrase generation as a sequence-to-sequence problem. Given some input sequence of tokens $x_{1:n} = (x_1,x_2,x_3,\dots,x_n)$ as the input query, we want a model to paraphrase the input into another sequence $y_{1:m} = (y_1,y_2,y_3,\dots,y_m)$ serving as the output. 

In our experiments, we mainly explore GPT-based models which are pre-trained on autoregressive language modeling or auto-completion. Thus, we frame our problem similar to an auto-completion task such that the downstream task remains similar to the pre-training task (potentially making the transfer of pre-trained knowledge easier). 

To frame paraphrasing as an auto-completion task for an autoregressive language model, we can design an \textbf{input prompt} such as ``Input: $x_1,x_2,x_3,\dots,x_n$ \textbackslash n Paraphrase: ". The task of the model is to auto-complete the given prompt. The completed sequence will look like: ``Input: $x_1,x_2,x_3,\dots,x_n$ \textbackslash n Paraphrase: $y_1,y_2,y_3,\dots,y_m$". We treat the generated sequence $y_{1:m}$ as the paraphrase of the original input $x_{1:n}$. 

In this overall sequence, ``input: " is the \textbf{prompt prefix}, ``\textbackslash n Paraphrase: " is the \textbf{prompt infix}, ``$x_1,x_2,x_3,\dots,x_n$" is the input, and ``$y_1,y_2,y_3,\dots,y_m$" is the paraphrased output. The prefix and infix together forms the prompt template: ``Input: \rule{1cm}{0.15mm}\textbackslash n Paraphrase: \rule{1cm}{0.15mm}". 

We can further generalize the input prompt format by formalizing a generic sequence of tokens denoting the prompt prefix as $p_{1:s} = (p_1,p_2,p_3,\dots,p_s)$ and a generic sequence of tokens denoting the prompt infix as $q_{1:t} = (q_1,q_2,q_3,\dots,q_t)$. Overall the generalized input prompt will be: ``$p_1,p_2,p_3,\dots,p_s, x_1,x_2,x_3,\dots,x_n,  q_1,q_2,q_3,\dots,q_t$". %During prompt tuning, the prefix sequence and the infix sequence can contain only virtual tokens which do not correspond to any natural language tokens or their corresponding embeddings. 
\section{Baselines}
\label{sec:baselines}
In this paper, we explore the following methods to adapt large pre-trained language models. Particularly, we use GPT-based models to adapt for paraphrase generation. 

\subsection{Fine Tuning}
Fine Tuning (FT) is the standard method of adapting a pre-trained model. In this strategy all the pre-trained parameters undergo gradient updates in some downstream tasks (in this case, paraphrase generation). In our implementation, we manually design an input prompt as ``Input: $x_1,x_2,x_3,\dots,x_n$ \textbackslash n Paraphrase: " where $x_{1:n} = x_1,x_2,x_3,\dots,x_n$ is the input as defined in the previous section. We trained the model in the auto-completion framework as discussed in the previous section.

\subsection{Adapter Tuning (AT)}
AT introduces some new sublayers (i.e., adapter layers) acting as low-rank bottlenecks within each Transformer layer. Generally, instead of tuning all parameters of a pre-trained model, AT focuses on tuning mainly the adapter layers. Following \citet{houlsby2019parameter}, we only tune the adapter layer parameters and the layer normalization parameters during training. We train the model with the same input prompt format and the same framework as used in fine tuning.

\subsection{Low Rank Adaptation (LoRA)}
Given a pre-trained matrix $W_x \in \mathbb{R}^{d \times k}$, LoRA \cite{hu2021lora} constrains the update of $W_x$ through a low-rank matrix $W_{\delta} \in \mathbb{R}^{d \times k}$.
More precisely, instead of directly updating $W_x$, LoRA first reformulates $W_x$ to $W'_x = W_x + W_{\delta}$, and then only updates the parameters involved in the construction of $W_{\delta}$. $W_{\delta}$ itself is constructed through the multiplication of two matrices $B \in \mathbb{R}^{d \times r}$ and $A \in \mathbb{R}^{r \times k}$ as: $W_{\delta} = BA$. We maintain $W_{\delta}$ as a low-rank matrix by keeping $r \ll min(d,k)$. Thus instead of tuning $dk$ parameters we only need to tune $r\cdot(d+k)$ parameters. %which can be significantly less when $r \ll min(d,k)$. 
Similar to \citet{hu2021lora}, we only apply LoRA to the query transformation and value transformation matrices in the multi-head attention sublayers. We train the model in the same framework as used in fine tuning. 

\subsection{Prompt Tuning}
Large-scale pre-trained language models can already contain substantial implicit knowledge for several natural language tasks before further fine tuning. For example, \citet{radford2019language} show that language models can serve as unsupervised multi-task learners. They show that using task-specific natural language prompts enable the pre-trained language models to do specific tasks like translation or summarization without further training. %In a similar vein, \citet{brown2020language} show that by providing input-output examples of a task in the input prompt itself, the model can learn the task in a few-shot manner. These works show the potential of simply using natural language prompts (prefix and infix sequences) along with the input text to guide the model for a desired task. 
However, manually designing prompt templates is not ideal because it requires human involvement and the ideal prompt may not correspond to human intuitions. Instead recent work \cite{shin2020autoprompt, li2021prefix, lester2021power} focuses on automatically tuning the prompt template itself. Moreover, some of these works also tend to tune only the prompt template parameters keeping all the pre-trained parameters frozen. As such, these methods also serve as lightweight parameter-efficient options to adapt large language models. %Our approach of prompt tuning is similar to the work of \citet{lester2021power}, where only the prefix-infix token embeddings are tuned in a continuous space.

For our implementation of prompt tuning, we consider the generalized input prompt format formalization as discussed in the previous section: 
``$p_1,p_2,p_3,\dots,p_s, x_1,x_2,x_3,\dots,x_n,  q_1,q_2,q_3,\dots,q_t$". The initial representation after passing this input prompt sequence through the embedding layer can be presented as: ``$e(p_1),\dots,e(p_s), e(x_1),\dots,e(x_n),  e(q_1),\dots,e(q_t)$". Similar to \citet{lester2021power}, %we do not try to either manually design natural language prompt prefix-infix sequences or constrain the prefix-infix tokens to correspond to discrete tokens from the natural language vocabulary of the pre-trained model. Instead, following the cited models, 
we directly initialize and tune the prefix ($e(p_1),\dots,e(p_s)$) and infix embeddings ($e(q_1),\dots,e(q_t)$). When using prompt tuning alone, we only tune the prefix and infix embeddings (unique embedding vectors for each position) while keeping all other parameters frozen.  We train the model in the same auto-completion framework as used in fine tuning.

\subsection{LoRA + Prompt Tuning (LPT)}
In this approach, we apply both prompt tuning and LoRA at the same time. We tune both the prefix-infix parameters and the newly introduced LoRA parameters. Both LoRA and prompt tuning tune only a minor fraction of the total parameters. Thus, we can easily combine them without increasing the overall parameter count significantly. %We train the model in the same auto-completion framework as used in fine tuning. 

\section{Proposed Approaches}

In this section, we introduce our proposed approaches: Retrieval Augmented Prompt Tuning (RAPT) and Novelty Conditioned Retrieval Augmented Prompt Tuning (NC-RAPT). 

\subsection{Retrieval Augmented Prompt Tuning (RAPT)}

In this section, we introduce our first proposed approach, RAPT, which is a synergy between an automated example retrieval strategy and an example-augmented prompt tuning strategy. We discuss both of them below. Like LPT, we also add LoRA because it does not add too many parameters (Table \ref{table:parameters}) and allows more tuning flexibility. 

\subsubsection{Example-Augmented Prompts:}
\citet{brown2020language} show that large-scale pre-trained auto-regressive models like GPT3 can locate or learn to perform a task from only a few input-output pairs exemplifying the task given in the input prompt context without further gradient updates. For example, let us say we have an input text sequence $X$ $($where $X = x_{1:n})$ that we want to paraphrase, and we have two exemplar paraphrase pairs $(X_1,Y_1)$ and $(X_2,Y_2)$. As such the input prompt to a GPT3-like model can be of the form:
\begin{equation}
    Z = [p_{1:s}; X_2; q_{1:t}; Y_2; p_{1:s}; X_1; q_{1:t}; Y_1; p_{1:s}; X; q_{1:t}],
    \label{naive_prompt_design}
\end{equation}
where $Z$ is the full input prompt sequence, $p_{1:s}$ is the prefix sequence for a sample input, and $q_{1:t}$ is the infix sequence marking the end of the sample input and the start of the sample output. $[.;.]$ denotes concatenation. 

We hypothesize that similarly augmenting the input prompt with task exemplars can be still beneficial for smaller pre-trained language models (e.g., GPT2). \citet{hu2021lora} show that full data fine tuning, LoRA, or prompt tuning on GPT3 can still outperform few-shot approaches.
Thus, instead of applying and exploring example-augmented prompts in few shot settings, we use example-augmented prompts in our full data prompt tuning setup.
That is, we can treat the embeddings of $p_{1:s}$ and $q_{1:t}$ (the prefix and infix embedding parameters as used in our prompt tuning setup as discussed before) as freely trainable parameters.
This strategy removes the need for manually designing prefix and infix sequences to best describe the task and delimit the different example inputs and outputs.

However, simply using the prompt design format as in Eq. \ref{naive_prompt_design} is not without its problems. During prompt tuning the prefix length ($k$) and/or the infix length ($l$) can be quite large (e.g., ~$250$). Since the prefix and infix are repeated for every example in Eq. \ref{naive_prompt_design}, the total length of the input prompt can increase significantly. Long sequences can slow down training and cause memory issues. %or even quickly run out of the maximum boundary if learnable position embeddings are involved. 
To solve this issue we revise the prompt design format in Eq. \ref{naive_prompt_design} as: 
\begin{equation}
    Z = [d_{1:m}; p_{1:s}; X_2; q_{1:t}; Y_2; p_{1:s}; X_1; q_{1:t}; Y_1; p_{1:s}; X; q_{1:t}].
    \label{less_naive_prompt_design}
\end{equation}
In this format, $d_{1:m}$ serves as a global prefix serving as a sort of task description. We try to maintain $s,t \ll m$ such that most of the tunable prompt embedding parameters are concentrated on the emebddings of $d_{1:m}$, allowing us to keep $s$ and $t$ small. Thus, given the small $s$ and $t$, the overall input prompt sequence length does not increase as much when $p_{1:s}$ and $q_{1:t}$ are repeated. 

\subsubsection{Example Retrieval:}
Above, we described how to design prompts to incorporate input-output examples of a task serving as task exemplars. However, the question remains open: \textit{where do we get the task examples?} We can always try to construct the task examples manually, but that would again require human involvement. %Additionally, the model can be sensitive to differences in task examples, and it is not always intuitively clear how to best construct examples for a task. 

Instead, we consider an automated method. We hypothesize that given an input sequence $X$, input-output example pairs that are most similar to $X$ would be the most suitable in serving as examples to guide paraphrasing of $X$. Thus, we use a k-Nearest-Neighbor (kNN) algorithm to retrieve the top-k example pairs that have the most semantically similar (to $X$) inputs from the training dataset. 

More precisely, we first encode each input text in the training data into a single sentence vector using sentence-transformers\footnote{\url{https://www.sbert.net/docs/pretrained_models.html}} \cite{reimers2019sentence-bert} (we use \verb+paraphrase-mpnet-base-v2+ as the pre-trained weights).
Similarly, we also encode any given input $X$ that we want to paraphrase. %Let us say $T(X)$ represents the encoded sentence embedding of any given sequence $X$. We then compute cosine similarity scores between $T(X)$ and the encoding $T(\widehat{X})$ of any input $\widehat{X}$ from the training dataset. %Let us say $C(T(X),T(\widehat{X}))$ denotes the cosine similarity score between $T(X)$ and $T(\widehat{X})$. 
We select top k examples $((X_1, Y_1), (X_2, Y_2),\dots, (X_k,Y_k))$ from the training set based on the cosine similarity scores between the embeddings of the training examples and that of the given input $X$. %$C(T(X), T(X_i))$ (where $i \in \{1,2,\dots,k\}$) ignoring any case where $X = X_i$. 
We add the retrieved examples to the prompt (as in Eq. \ref{less_naive_prompt_design}) in the ascending order of their cosine similarity scores with the input $X$. In practice, we set $k=2$ given the disadvantages of long sequences that we discussed. 
%\subsubsection{RAPT:} 

\subsection{Novelty Conditioned RAPT (NC-RAPT)}
%If we want to make non-trivial paraphrases we also want them to be novel. %Otherwise, we can make trivial paraphrases by substituting single words with synonyms using simpler algorithms. 
While there are prior works \cite{gupta2018deep, kumar2019submodular} that have focused on improving the novelty of paraphrase generation, there are not many works focusing on controlling the level of novelty. In this work, we use a simple model-agnostic framework to control the novelty level of the generation. This framework allows us at least two benefits.
\begin{enumerate}
    \item The framework puts more control over the user in deciding the trade-offs they want in increasing the novelty of paraphrasing. 
    \item The framework can also enforce a model to generate highly novel paraphrases and thus compete with prior novelty-enhancing methods. 
\end{enumerate}
Our framework requires two broad steps. In the first step, we classify each sample in the dataset according to the novelty of the ground truth paraphrase. %We consider that the edit-distance between the input and the ground truth output can be treated as the measure of the novelty of ground truth with respect to the input. 
We use Translation Edit Rate (TER) \cite{olive2005gale, snover2006study} between the input sequence and the ground truth sequence to quantify the novelty of the ground truth. %TER can approximately compute the token-level edit distance between two given texts. 
We use three classes of novelty levels: \verb+High+, \verb+Medium+, and \verb+Low+. We classify each sample among the three novelty level classes based on the TER values between the input and the ground truth. If the TER is $\ge 0.4$, we classify the sample as \verb+High+. If the TER is $> 0.2$ and $< 0.4$, we classify the sample as \verb+Medium+. If the TER is $\le 0.2$, we classify the sample as \verb+Low+. The thresholds were chosen intuitively based on qualitative analysis. 

In the second step, during training, we use different prefix and infix embeddings for different novelty classes. Let us say $c_{(X,Y)}$ denotes the novelty-class of an input-output paraphrase-pair $(X,Y)$, $p^c_{1:s}$ denotes the prefix sequence for novelty class $c$, and $q^c_{1:t}$ denotes the infix sequence for novelty class $c$. Given these notations, during training, we reformulate Eq. \ref{less_naive_prompt_design} for novelty-conditioned generation as:
\begin{equation}
\begin{split}
    Z = [&d_{1:m};p^{c_{(X_2,Y_2)}}_{1:s};X_2;q^{c_{(X_2,Y_2)}}_{1:t};Y_2;\\
         &p^{c_{(X_1,Y_1)}}_{1:s};X_1; q^{c_{(X_1,Y_1)}}_{1:t};Y_1;p^{c_{(X,Y)}}_{1:s};X; q^{c_{(X,Y)}}_{1:t}].
    \label{novelty_conditioned_less_naive_prompt_design}
\end{split}
\end{equation}
Here $Y$ is the ground truth paraphrase for $X$. We keep the same global prefix sequence $d_{1:m}$ for all novelty classes to save parameters. As can be understood from Eq. \ref{novelty_conditioned_less_naive_prompt_design}, the model learns from a ground truth with novelty class $c$ only when the corresponding prefix ($p^c_{1:s}$) and infix ($q^c_{1:t}$) sequences for the same novelty class $c$ are used. As such, during inference, the model learns to generate a paraphrase of novelty class $c$ if the prefix ($p^c_{1:s}$) and infix ($q^c_{1:t}$) sequences for novelty class $c$ is used for the input. During inference, to generate a paraphrase of a given input $X$ with novelty class $c$, we simply use $p^{c}_{1:s}$ instead of $p^{c_{(X,Y)}}_{1:s}$ and $q^{c}_{1:t}$ instead of $q^{c_{(X,Y)}}_{1:t}$ in Eq. \ref{novelty_conditioned_less_naive_prompt_design}.
\begin{comment}
\begin{equation}
\begin{split}
    Z = [&d_{1:m};p^{c_{(X_2,Y_2)}}_{1:s};X_2;q^{c_{(X_2,Y_2)}}_{1:t};Y_2;\\       
         &p^{c_{(X_1,Y_1)}}_{1:s};X_1; q^{c_{(X_1,Y_1)}}_{1:t};Y_1;\\
         &p^{c}_{1:s};X; q^{c}_{1:t}].
    \label{novelty_conditioned_less_naive_prompt_design_inference}
\end{split}
\end{equation}
\end{comment}
NC-RAPT is built upon RAPT. So every other details are the same as for RAPT. We show some example model generations in Table \ref{table:samples}. In Table \ref{table:parameters}, we show the total trainable parameters for all the adaptation methods that we try over GPT2 medium and GPT2 large. 

%\todo{jishnu}{show generation examples for different levels of diversity}
%We also explored the option of retrieving top-k examples from the same novelty class as used for the given input, but it did not benefit us significantly. 

\begin{table}[t]
\centering
\def\arraystretch{1.5}
\begin{tabularx}{\linewidth}{lX} 
\toprule
\textbf{Input} & \emph{Why do all of my questions get marked as needing improvement}\\
\midrule
\textbf{LPT} & why do my questions get marked as needing improvement\\
\midrule
\textbf{RAPT} & why are my questions still being marked as needing improvement\\
\midrule
\textbf{NC-RAPT (High)} & why is my question being marked as needs to be improved\\
\midrule
\textbf{NC-RAPT (Low)} & why do all of my questions get marked as needing improvement\\
\bottomrule
\end{tabularx}
\caption{Examples of generated paraphrases by different models from Quora Question Pairs dataset.} 
\label{table:samples}
\end{table}

\begin{table}[t]
\centering
\def\arraystretch{1.2}
\begin{tabularx}{\linewidth}{Xrr} 
\toprule
 & \multicolumn{2}{c}{\textbf{Number of Parameters}}\\
\cmidrule(lr){2-3} 
\textbf{Model} & \textbf{GPT2 Medium} &  \textbf{GPT2 Large}\\
\midrule
Fine Tuning &   354,823,168  & 774,030,080\\
Adapter Tuning &   25,303,040  & 47,437,312\\
LoRA Tuning &   786,432  & 1,474,560\\
Prompt Tuning &   270,336  & 337,920\\
LPT  &   1,056,768 & 1,812,480\\
RAPT & 1,056,768 & 1,812,480\\
NC-RAPT & 1,089,536 & 1,853,440\\
\bottomrule
\end{tabularx}
\caption{Comparison on the number of trainable parameters for different adaptation methods.} 
\label{table:parameters}
\end{table}

\section{Experiments}
%In this section, we first introduce the dataset used in our experiments. Second, we discuss about our evaluation methods. Next, we share some experimental details. 
%Finally, we share and discuss our results. %We share other experimental details and hyperparameters in the Appendix \todo{jishnu}{to add reference and appendix}.

\subsection{Dataset}
We use four datasets for our experiments as discussed below. Details of dataset split sizes are presented in Table \ref{table:dataset details}. 

\begin{table}[h]
\centering
\def\arraystretch{1.2}
\begin{tabular}{  l  r  r  r } 
\toprule
\textbf{Dataset Name} & \textbf{Training} & \textbf{Validation} & \textbf{Test}\\
\midrule
QQP 50K& 46,000 & 4,000 & 4,000\\
QQP 140K& 134,206 & 5255 & 5255\\
MSRPC & 2,203 & 550 & 1,147\\
ParaSCI ACL & 28,883 & 2753 & 2,345\\
\bottomrule
\end{tabular}
\caption{Dataset split sizes} 
\label{table:dataset details}
\end{table}
\begin{itemize}
    \item \textbf{Quora Question Pairs 50K split (QQP 50K)}\footnote{\url{https://quoradata.quora.com/First-Quora-Dataset-Release-Question-Pairs}}: Quora Question Paris (QQP) is a paraphrase detection dataset. We only use the true paraphrase pairs. We use the 50K dataset split as used in \citet{gupta2018deep}.\footnote{\url{https://github.com/arvind385801/paraphraseGen}}  
    \item \textbf{Quora Question Pairs 140K split (QQP 140K)}: We also use QQP with a different split size (QQP 140K) as used by \citet{hosking2021factorising}. \footnote{\url{https://github.com/tomhosking/separator}}
    \item \textbf{Microsoft Research Paraphrase Corpus (MSRPC)}: MSRPC \cite{dolan2004unsupervised} is another paraphrase detection corpus. We use only the true paraphrase pairs for paraphrase generation. We use the official train-test split.
    %\item \textbf{ComQA} \cite{abujabal-etal-2019-comqa} is a large dataset of real user questions coming from the WikiAnswers community QA platform. This dataset groups questions into parapharse clusters through a large crowdsourcing effort.
    \item \textbf{ParaSCI-ACL}: ParaSCI-ACL \cite{dong2021parasci} is a paraphrase generation dataset in the scientific domain. We use the official split.\footnote{\url{https://github.com/dqxiu/ParaSCI}} 
\end{itemize}

\begin{table*}[t]
\small
\centering
\def\arraystretch{1.2}
\begin{tabular}{  l  r R{0.8cm} R{0.8cm} r r r | r R{0.8cm} R{0.8cm} r r r } 
\toprule
& \textbf{BERT} & \textbf{Self-TER} & \textbf{Self-BLEU} &\textbf{BLEU} &\textbf{iBLEU} & \textbf{SARI} & \textbf{BERT} & \textbf{Self-TER} & \textbf{Self-BLEU} &\textbf{BLEU} &\textbf{iBLEU} & \textbf{SARI}\\
\hline
%& \multicolumn{6}{c|}{\textbf{Dataset: QQP 140K}} & \multicolumn{6}{c}{\textbf{Dataset: QQP 50K}}\\
\textbf{Method} & \multicolumn{6}{c}{\textbf{Dataset: QQP 140K}}&\multicolumn{6}{c}{\textbf{Dataset: QQP 50K}}\\
\cmidrule{2-7}
\cmidrule{8-13}
%\hline
Copy & 100.0 & 0.00 & 100.0 & 32.78 & -7.05 & 14.98 & 100.0  & 0.00 & 100.0 & 30.36 & -8.75 & 14.44\\
Ground Truth & 89.05  & 54.31 & 30.98 & 100.0 & 60.71 & 83.88 & 88.53  & 56.31 & 30.34 & 100.0 & 60.90 & 87.51\\
\hline
\multicolumn{13}{l}{GPT2 Baselines}\\
\hline
Fine Tuning & 94.39 & 29.57 & 58.80 & 33.26 & 5.64 & 38.02 & 92.76  & 32.97 & 54.40 & 29.77 & 4.52 & 38.96\\
Adapter Tuning & 93.77 & 35.89 & 56.75 & 30.20 & 4.12 & 36.64 & 91.90 & 36.68 & 52.89 & 27.90 & 3.66 & 37.61\\
LoRA & 92.92 & 40.87 & 50.16 & 27.74 & 4.37 & 37.89 & 92.03  & 40.09 & 48.98 & 26.68 & 3.98 & 38.87\\
Prompt Tuning & 94.81 & 25.96 & 59.94 & 31.56 & 4.11 & 36.45 & 91.79  & 34.23 & 47.63 & 27.34 & 4.85 & 40.49\\
LPT & 94.99 & 27.40 & 59.16 & 33.21 & 5.50 & 38.67 & 92.62  & 34.84 & 47.79 & 28.70 & 5.75 & 41.58\\
\hline
\multicolumn{13}{l}{Ours}\\
\hline
RAPT & 87.88 & 50.27 & 31.78 & 34.09 & 14.33 & 42.40 & 90.04  & 44.99 & 38.39 & \textbf{31.61} & 10.61 & 43.91\\
NC-RAPT (High) & 83.89 & \textbf{62.13} & \textbf{19.55} & 29.88 & \textbf{15.05} & 42.72 & 86.98  & \textbf{58.27} & \textbf{23.15} & 25.12 & \textbf{10.64} & 43.23\\
NC-RAPT (Med) & 89.84 & 43.47 & 37.77 & 36.47 & 14.19 & \textbf{44.67} & 92.25  & 37.40 & 43.39 & 30.22 & 8.14 & \textbf{45.16}\\
NC-RAPT (Low) & \textbf{96.97} & 16.22 & 76.60 & \textbf{38.36} & 3.87 & 31.14 & \textbf{96.78}  & 18.45 & 69.16 & 31.35 & 1.20 & 40.04\\
%RANCoPT (Ground Truth) & 1.09M & 87.34 & 51.43 & 32.19 & 38.44 & 17.25 & 45.01 & 50.26\\
\midrule
%& \multicolumn{6}{c|}{\textbf{Dataset: ParaSCI ACL}} & \multicolumn{6}{c}{\textbf{Dataset: MSRPC}}\\
%\hline
\textbf{Method} & \multicolumn{6}{c}{\textbf{Dataset: ParaSCI ACL}}&\multicolumn{6}{c}{\textbf{Dataset: MSRPC}}\\
\cmidrule{2-7}
\cmidrule{8-13}
Copy & 100.00 & 0.00 & 100.0 & 35.66 & -5.04 & 15.01 & 100.0 & 0.00 & 100.0 & 47.75 & 3.43 & 19.91 \\
Ground Truth & 78.55 & 61.22 & 35.76 & 100.00 & 59.27 & 90.57 & 86.15 & 48.03 & 47.77 & 100.0 & 55.67 & 96.72\\
\hline
\multicolumn{7}{l}{GPT2 Baselines}\\
\hline
Fine Tuning & 89.94 & 28.56 & 72.34 & 33.32 & 1.62 & 33.66 & 97.35 & 9.09 & 89.63 & 45.09 & 4.67 & 29.78 \\
Adapter Tuning & 87.47 & 35.17 & 64.5 & 29.79 & 1.50 & 34.00 & 29.54 & \textbf{73.15} & \textbf{8.85} & 4.23 & 0.31 & 21.26\\
LoRA & 92.43 & 24.28 & 74.37 & 31.99 & 0.08 & 33.22 & 96.85 & 10.7 & 86.95 & 44.57 & 5.12 & 32.81\\
Prompt Tuning & 91.05 & 23.48 & 71.18 & 29.93 & -0.40 & 31.88 & 95.76 & 14.59 & 80.68 & 42.13 & 5.29 & 34.51\\
LPT & 92.83 & 19.6 & 77.06 & 33.08 & 0.04 & 31.64 & 95.57 & 15.05 & 80.21 & 41.7 & 5.13 & 34.99\\
\hline
\multicolumn{7}{l}{Ours}\\
\hline
RAPT & 84.86 & 41.48 & 55.37 & 35.34 & 8.13 & 39.00 & 94.54 & 16.93 & 79.84 & 45.08 & 7.60 & 35.58\\
NC-RAPT (High) & 79.44 & \textbf{55.88} & \textbf{39.29} & 30.32 & \textbf{9.44} & \textbf{41.30} & 91.79 & 24.78 & 68.64 & 40.42 & \textbf{7.70} & \textbf{38.6} \\
NC-RAPT (Med) & 89.49 & 31.17 & 63.94 & 36.30 & 6.23 & 40.44 & 96.56 & 11.57 & 85.42 & 46.14 & 6.67 & 32.88\\
NC-RAPT (Low) & \textbf{94.62} & 17.17 & 79.37 & \textbf{39.16} & 3.60 & 35.54 & \textbf{97.74} & 7.72 & 90.22 & \textbf{47.65} & 6.29 & 30.18 \\
%RANCoPT (Ground Truth) & 1.09M & 82.10 & 49.02 & 46.92 & 36.59 & 11.54 & 43.23 & 46.47\\
\bottomrule
\end{tabular}
%\vspace{+.5em}
\caption{Comparison of different approachs to adapt \textbf{GPT2 Medium} for paraphrase generation on different datasets. We bold the best results for each dataset excluding Copy and Ground Truth} 
\label{table:main results}
\end{table*}
\begin{table*}[t]
\small
\centering
\def\arraystretch{1.2}
\begin{tabular}{  l r R{0.8cm} R{0.8cm} r r r | r R{0.8cm} R{0.8cm} r r r } 
\toprule
& \textbf{BERT} & \textbf{Self-TER} & \textbf{Self-BLEU} &\textbf{BLEU} &\textbf{iBLEU} & \textbf{SARI} & \textbf{BERT} & \textbf{Self-TER} & \textbf{Self-BLEU} &\textbf{BLEU} &\textbf{iBLEU} & \textbf{SARI}\\
%& \multicolumn{6}{c|}{\textbf{Dataset: QQP 50K}} & \multicolumn{6}{c}{\textbf{Dataset: MSRPC}}\\
\midrule
\textbf{Method} & \multicolumn{6}{c}{\textbf{Dataset: QQP 50K}}&\multicolumn{6}{c}{\textbf{Dataset: MSRPC}}\\
\cmidrule{2-7}
\cmidrule{8-13}

\multicolumn{13}{l}{GPT2 Baselines}\\
\hline
Fine Tuning & 92.49 & 34.56 & 50.96 & 29.82 & 5.59 & 40.13 & 95.55 & 14.0 & 84.10 & 43.96 & 5.54 & 34.49\\
Adapter Tuning & 92.35 & 36.51 & 49.55 & 29.16 & 5.55 & 40.04 & 93.91 & 13.84 & 81.81 & 42.82 & 5.43 & 33.96\\
LoRA & 92.98 & 33.47 & 52.96 & 29.67 & 4.88 & 39.81 & 96.80 & 11.31 & 86.11 & 44.88 & 5.58 & 33.62\\
Prompt Tuning & 93.51 & 29.94 & 54.82 & 29.62 & 4.29 & 39.39 & 96.73 & 10.82 & 86.46 & 44.94 & 5.52 & 32.82\\
LPT & 92.94 & 33.63 & 50.32 & 29.60 & 5.63 & 40.91 & 97.20 & 8.55 & 88.26 & 45.52 & 5.38 & 32.15\\
\hline
\multicolumn{13}{l}{Ours}\\
\hline
RAPT & 90.56 & 42.83 & 41.11 & 31.16 & 9.48 & 43.27 & 96.50 & 11.34 & 86.06 & 47.80 & 7.64 & 34.57\\
NC-RAPT (High) & 87.13 & \textbf{56.45} & \textbf{25.04} & 26.33 & \textbf{10.92} & 43.25 & 92.37 & \textbf{23.9} & \textbf{69.33} & 41.40 & \textbf{8.18} & \textbf{39.50}\\
NC-RAPT (Med) & 92.11 & 36.15 & 45.25 & 31.75 & 8.65 & \textbf{44.99} & 97.56 & 8.50 & 88.90 & 47.34 & 6.47 & 30.89\\
NC-RAPT (Low) & \textbf{96.69} & 17.43 & 71.29 & \textbf{32.03} & 1.03 & 38.96 & \textbf{98.85} & 4.32 & 94.32 & \textbf{48.86} & 5.90 & 27.42\\
%RANCoPT (Ground Truth) & & 89.21 & 48.4 & 34.33 & 33.49 & 13.14 & 46.31 & 49.27\\
\bottomrule
\hline
\end{tabular}
%\vspace{+.5em}
\caption{Comparison of different approaches to adapt \textbf{GPT2 Large} for paraphrase generation on different datasets. We bold the best results for each dataset excluding Copy and Ground Truth.} 
\label{table:large gpt results}
\end{table*}

\subsection{Evaluation Metrics}
\label{sec:metrics}
We use different evaluation metrics to account for different aspects of paraphrase quality as below:
\begin{itemize}
\item \textbf{BLEU}: Like most prior work, we use BLEU4 \cite{papineni2002bleu} to measure similarity between prediction and ground truths. 

\item \textbf{Self-BLEU}: BLEU by itself does not explicitly measure for the novelty of the generated paraphrase. Often, it is possible to achieve high BLEU in paraphrase generation by simply copying the original input. Thus, we also use self-BLUE4, i.e. BLEU4 between the input and the prediction to account for the novelty of the prediction. Low Self-BLEU implies high novelty.
\item \textbf{Self-TER}: To check for the novelty of paraphrases beyond simply checking n-gram overlaps as in Self-BLEU, we also measure the \textbf{T}ranslation \textbf{E}dit \textbf{R}ate (TER) \cite{snover2006study}  between the input and the prediction. High self-TER implies high novelty. 

\item \textbf{BERT}: We also want to account for the semantic fidelity of the generated texts. A paraphrase must retain the main semantic content of the input. To measure this, we use cosine-similarity between the sentence encodings of the input and the prediction. For sentence encoding we use Sentence-BERT \cite{reimers2019sentence-bert} along with \verb+paraphrase-mpnet-base-v2+\footnote{\url{https://www.sbert.net/docs/pretrained_models.html}} as the pre-trained weights.

\item \textbf{iBLEU}: iBLEU \cite{sun2012joint} is designed to combine both BLEU and self-BLEU to measure the balance between closeness to ground truth and novelty:
\begin{equation}
    \text{iBLEU} = \alpha \cdot \text{BLEU} - (1-\alpha)\cdot\text{self-BLEU}
\end{equation}
We use BLEU4 and self-BLEU4 with $\alpha$ being $0.7$ similar to \citet{hosking2021factorising}.

\item \textbf{SARI}: SARI \cite{xu2016optimizing} explicitly checks for the goodness of edits made on the input by the model in its predictions by comparing against the edits in the ground truth references. %was originally proposed as a metric for text-simplification but it has been used as a measure for paraphrase generation as well \cite{west2021reflective} . 
\citet{west2021reflective} show that it correlates well with human judgment for paraphrase generation. %SARI explicitly checks for the goodness of edits made on the input by the model in its predictions by comparing against the edits in the ground truth references.  
%\item \textbf{AVG}: As an overall score, we also report the average of all the above metrics exlcuding BLEU4 and self-BLEU4. We exclude them from the average because they are already included in iBLEU4 which is itself part of the average. 
    
\end{itemize}

\subsection{Experimental Details}
In the main experiments (Table \ref{table:main results}), we explore different adaptation methods on GPT2 medium. We share some results (Table \ref{table:large gpt results}) on GPT2 large too. Typically, during prompt tuning, we initialized the prompt prefix and infix token embeddings with random natural language token embeddings from GPT2 vocabulary. Besides the language model adaptation techniques, we also compare a copy-baseline that simply copies the input text and poses it as a paraphrase. The copy model checks how far we can go in scores like BLEU without changing a single word. We also try using the ground truths as predictions themselves to get a better sense of natural novelty and semantic fidelity of the ground truths in a particular dataset. We share other experimental details and hyperparameters in a later section.

%\begin{table}[t]
%\small
%\centering
%\def\arraystretch{1.2}
%\begin{tabularx}{\linewidth}{  X r r r } 
%\toprule
%\textbf{Method} & \textbf{Self-BLEU} &\textbf{BLEU} &\textbf{iBLEU}\\
%& \multicolumn{3}{c}{\textbf{Dataset: QQP 140K}} \\
%\cmidrule{2-4}
%\multicolumn{4}{l}{Prior Work}\\
%\midrule
%\textbf{Prior Work} & \multicolumn{3}{c}{\textbf{Dataset: QQP 140K}}\\
%\cmidrule{2-4}
%DiPS$^*$ & 32.45 & 18.47 & 3.19 \\
%SOW/REAP$^*$ & 24.19 & 12.64 & 1.59 \\
%LBoW$^*$ & 29.00 & 16.17 & 2.62 \\
%SEPARATOR$^*$ & \textbf{14.84} & 14.70 & 5.84 \\
%\midrule
%\multicolumn{4}{l}{Ours}\\
%\midrule
%\textbf{Ours} & \multicolumn{3}{c}{\textbf{Dataset: QQP 140K}}\\
%\cmidrule{2-4}
%RAPT & 31.78 & \textbf{34.09} & 14.33\\
%NC-RAPT (High) & 19.55 & 29.88 & \textbf{15.05}\\
%NC-RAPT (Med) & 37.77 & 36.47 & 14.19 \\
%NC-RAPT (Low) & 76.60 & \textbf{38.36} & 3.87\\
%\bottomrule
%\end{tabularx}
%\vspace{+.5em}
%\caption{Comparison with prior work. $^*$ indicates that the results are taken from \citet{hosking2021factorising}. We bold the best results. DiPS refers to the model proposed by \citet{kumar2019submodular}. SOW/REAP refers to the model proposed by \citet{goyal2020neural}. LBoW refers to the model proposed by \citet{fu2019paraphrase}. Separator refers to the model proposed by \citet{hosking2021factorising}. Our models are based on GPT2 medium.} 
%\label{table:prior work}
%\end{table}

\begin{table}[t]
\small
\centering
\def\arraystretch{1.2}
\begin{tabularx}{\linewidth}{  X r r r } 
\toprule
\textbf{Method} & \textbf{Self-BLEU} &\textbf{BLEU} &\textbf{iBLEU}\\
%& \multicolumn{3}{c}{\textbf{Dataset: QQP 140K}} \\
%\cmidrule{2-4}
%\multicolumn{4}{l}{Prior Work}\\
\midrule
\cmidrule{2-4}
DiPS$^*$ & 32.45 & 18.47 & 3.19 \\
SOW/REAP$^*$ & 24.19 & 12.64 & 1.59 \\
LBoW$^*$ & 29.00 & 16.17 & 2.62 \\
SEPARATOR$^*$ & \textbf{14.84} & 14.70 & 5.84 \\
%\midrule
%\multicolumn{4}{l}{Ours}\\
%\midrule
%\cmidrule{2-4}
RAPT & 31.78 & \textbf{34.09} & 14.33\\
NC-RAPT (High) & 19.55 & 29.88 & \textbf{15.05}\\
%NC-RAPT (Med) & 37.77 & 36.47 & 14.19 \\
%NC-RAPT (Low) & 76.60 & \textbf{38.36} & 3.87\\
\bottomrule
\end{tabularx}
%\vspace{+.5em}
\caption{Comparison with prior work on QQP 140K. $^*$ indicates that the results are taken from \citet{hosking2021factorising}. We bold the best results. DiPS refers to the model proposed by \citet{kumar2019submodular}. SOW/REAP refers to the model proposed by \citet{goyal2020neural}. LBoW refers to the model proposed by \citet{fu2019paraphrase}. Separator refers to the model proposed by \citet{hosking2021factorising}. Our models are based on GPT2 medium.} 
\label{table:prior work}
\end{table}

\subsection{Experimental Results}
%As can be seen in Table \ref{table:main results}, our proposed approaches (RAPT and NC-RAPT) generally substantially outperform the baselines on most metrics. Simply from using RAPT we get a major improvement in all the metrics except BERT. NC-RAPT behaves mostly as expected. We get high novelty (high self-TER, low self-BLEU) when using prefix and infix for novelty class \verb+high+; we get low novelty when conditioned on novelty class \verb+low+. We also observe a trade-off among novelty metrics, BLEU, and BERT in NC-RAPT. Conditioning on high novelty can reduce semantic fidelity (BERT) and BLEU compared to RAPT or NC-RAPT conditioned on \verb+medium+ or \verb+low+. However, the reduced semantic fidelity (BERT) score is still similar to or better than the same score as achieved by ground truth paraphrases (Ground Truth Method). The semantic fidelity (BERT) score of specifically \verb+high+ novelty class ground truth paraphrases should be even lower.  NC-RAPT conditioned on \verb+medium+ can sometimes gain better SARI scores whereas NC-RAPT conditioned on \verb+low+ can get the best BERT (semantic fidelity) and BLEU scores but at a significant cost to novelty. NC-RAPT allows the users to decide which trade-offs they are willing to take by choosing a specific novelty class. 
As shown in Table \ref{table:main results}, RAPT substantially outperforms the baselines on all metrics except BERT in general. Increased novelty may slightly hurt BERT-based semantic similarity score. %Conditioning on high novelty can reduce semantic fidelity (BERT) and BLEU compared to RAPT or NC-RAPT conditioned on \verb+medium+ or \verb+low+. However, the reduced semantic fidelity (BERT) score is still similar to or better than the same score as achieved by ground truth paraphrases (Ground Truth Method). The semantic fidelity (BERT) score of specifically \verb+high+ novelty class ground truth paraphrases should be even lower.  NC-RAPT conditioned on \verb+medium+ can sometimes gain better SARI scores whereas NC-RAPT conditioned on \verb+low+ can get the best BERT (semantic fidelity) and BLEU scores but at a significant cost to novelty. NC-RAPT allows the users to decide which trade-offs they are willing to take by choosing a specific novelty class. 

NC-RAPT behaves mostly as expected. We get high novelty (high self-TER, low self-BLEU) when using prefix and infix for novelty class \verb+high+; we get low novelty when conditioned on \verb+low+ novelty. We also observe a trade-off among novelty metrics, BLEU, and BERT in NC-RAPT.
Conditioning on \verb+high+ novelty increases novelty but can reduce semantic fidelity (BERT) and BLEU compared to other models. However, the reduced semantic fidelity (BERT) score is still similar to or better than that achieved by ground truth paraphrases (Ground Truth Method). NC-RAPT conditioned on \verb+high+ generally gets better novelty scores and iBLEU compared to RAPT and the baselines. NC-RAPT conditioned on \verb+medium+ can sometimes gain the best SARI scores whereas NC-RAPT conditioned on \verb+low+ can get the best BERT and BLEU scores but at a significant cost to novelty. NC-RAPT allows the users to decide which trade-offs they are willing to take by choosing a specific novelty class.

%Conditioning on low novelty can result in high semantic fidelity (BERT) and BLEU at a severe cost to novelty. 
The copy models can still achieve relatively high BLEU despite simply copying the original text. This phenomenon was also noted by \citet{mao2019polly}. This shows the limitation of relying on metrics like BLEU alone. 

In Table \ref{table:large gpt results}, we show that we still get fairly consistent results even when we use GPT2 large. %instead of GPT2 medium for our adaptation approaches.
In Table \ref{table:prior work}, we compare with some reported results of prior work (described in the table caption). Although SEPARATOR gets the highest novelty score (lowest self-BLEU), its BLEU score is quite low. Our proposed approaches based on GPT2 medium generally achieve a much better balance between BLEU and self-BLEU and thus, a substantially higher iBLEU.
%To demonstrate the effectiveness of the proposed approach, we compare the state-of-the-art approaches including fine tuning GPT-2 models and the diversity loss introduced in Section \ref{sec:methods}.

\begin{table*}[t]
\small
\centering
\def\arraystretch{1.2}
\begin{tabular}{  l  r R{0.8cm} R{0.8cm} r r  r | r R{0.8cm} R{0.8cm}  r  r  r } 
\toprule
& \textbf{BERT} & \textbf{Self-TER} & \textbf{Self-BLEU} &\textbf{BLEU} &\textbf{iBLEU} & \textbf{SARI} & \textbf{BERT} & \textbf{Self-TER} & \textbf{Self-BLEU} &\textbf{BLEU} &\textbf{iBLEU} & \textbf{SARI}\\
\hline
\textbf{Method} & \multicolumn{6}{c}{\textbf{Dataset: QQP 50K}}&\multicolumn{6}{c}{\textbf{Dataset: ParaSCI ACL}}\\
\cmidrule{2-7}
\cmidrule{8-13}
%& \multicolumn{6}{c|}{\textbf{Dataset: QQP 50K}} & \multicolumn{6}{c}{\textbf{Dataset: ParaSCI ACL}}\\
%\hline
RAPT (kNN) & 90.04 & \textbf{44.99} & \textbf{38.39} & \textbf{31.61} & \textbf{10.61} & \textbf{43.91} & 84.86 & \textbf{41.48} & \textbf{55.37} & \textbf{35.34} & \textbf{8.13} & \textbf{39.00}\\
RAPT (random) & \textbf{92.48} & 35.39 & 48.11 & 29.50 & 6.22 & 41.65 & \textbf{92.45} & 20.76 & 75.89 & 32.82 & 0.21 & 31.98 \\
\bottomrule
\end{tabular}
%\vspace{+.5em}
\caption{Comparison of kNN-based RAPT and RAPT with random retrieval. The results are based on adapting GPT2 medium.} 
\label{table:rapt ablation}
\end{table*}

\begin{table}[t]
\small
\centering
\def\arraystretch{1.2}
\begin{tabular}{  l r R{0.8cm} r r} 
\toprule
\textbf{Method}  & \textbf{BERT} & \textbf{Self-TER} &\textbf{iBLEU} & \textbf{SARI}\\
%\midrule
%\textbf{Method} & \multicolumn{4}{c}{\textbf{Dataset: QQP 50K}}\\
\cmidrule{2-5}
Prompt Tuning & 91.79 & 34.23 & \textbf{4.85} & \textbf{40.49}\\
Prompt Tuning Random & 92.22 & 33.56 & 4.42 & 39.70\\
P-Tuning & \textbf{93.90} & 27.74 & 2.66 & 37.33\\
Prefix Tuning & 89.25 & 43.98 & 4.34 & 38.69\\
Prefix-layer Tuning & 88.64 & \textbf{47.62} & 4.63 & 38.58\\
\bottomrule
\end{tabular}
%\vspace{+.5em}
\caption{Comparison of Prompt Tuning variants when adapting GPT2 medium on QQP 50K.} 
\label{table:pt variants}
\end{table}

\subsection{Analysis}
\textbf{Random Retrieval Ablation:} When using RAPT we use a kNN based example retrieval as  discussed before. As an ablation, we try random retrieval instead of a kNN-based approach. As shown in Table \ref{table:rapt ablation}, retrieving semantically similar examples using kNN gives us a much better performance. These results support our hypothesis that semantically similar example pairs can better guide paraphrase generation. 

However, as a collolary these results may also suggest that RAPT-based methods would not perform as well on a testing dataset where the examples are semantically distant from anything in the training dataset. This can be one limitation of the method. 

\noindent \textbf{Prompt Tuning Variants:} We also experimented with a few different variants of prompt tuning for paraphrase generation. We reported the results in Table \ref{table:pt variants}. Here, Prompt Tuning \cite{lester2021power} is the method that we use in our main experiments. Prompt Tuning Random is the same method but with random initialization of the prefix-infix embeddings instead of random selection from vocabulary embeddings. P-Tuning is similar to the method proposed by \citet{liu2021gpt}. It uses a BiLSTM to associate the prefix and infix tokens. Prefix Tuning is similar to the method proposed by \citet{li2021prefix}. It uses the same prefix, infix (transfromed by an MLP layer) in every GPT2 layer. Prefix-layer Tuning was used by \citet{hu2021lora}. It initializes and trains prefix-infix parameters for every layer instead of just the embedding layer as in simple prompt tuning \cite{lester2021power}. In the results, we see that prompt tuning gets the best iBLEU and SARI scores. Although Prefix and Prefix-layer Tuning increase the novelty it does so at the cost of other metrics. Thus, we chose prompt tuning for our main experiments. 

\subsection{Hyperparameter Details}

\begin{table*}[t]
\centering
\def\arraystretch{1.2}
\begin{tabular}{  l r  r  r  r} 
\toprule
& \textbf{Learning Rate} & \textbf{Adapter Bottleneck} & \textbf{prefix length} & \textbf{Prompt Dimension (b)}\\
\midrule
\textbf{Method} & \multicolumn{4}{c}{\textbf{Dataset: QQP 50K}}\\
\cmidrule{2-5}
%\hline
Fine Tuning & $5e-5$ & --- & --- & ---\\
Adapter Tuning & $1e-4$ & $512$ & --- & ---\\
LoRA & $1e-3$ & --- & --- & ---\\
Prompt Tuning & $0.1$ &---& $256$ & ---\\
Prompt Tuning Random & $0.1$ &---& $256$ & ---\\
P-Tuning & $0.01$ &---& $256$ & $d$\\
Prefix Tuning & $1e-4$ & --- & $256$ & $d$\\
Prefix-layer Tuning & $0.01$ & --- & $64$ & ---\\
\bottomrule
\end{tabular}
\caption{Selected Hyperparameters. $d$ is the embedding dimension of the involved pre-trained language model.} 
\label{table:hyperparemeters}
\end{table*}

We tune the hyperparameters on QQP 50K with GPT2 medium for all the approaches. We then use the same hyperparameters for other datasets and GPT2 large. We use random search with a maximum of $50$ trials and $3$ epochs per trial, and choose the hyperparameters based on validation loss. 
For all the approaches, we search the learning rate within $\{0.1, 0.01, 1e-3, 1e-4, 5e-5\}$. 
For adapter tuning, we search the adapter bottleneck hidden state dimension within $\{128, 256, 512\}$. 

For LPT, we tune the learning rate for LoRA paramters and prompt tuning parameters separately (we use different learning rates to tune LoRA and prompt template embeddings). For LoRA, LPT, RAPT, and NC-RAPT (all approached involving LoRA), we fix $r$ (matrix rank) as $8$ because it worked well in \citet{hu2021lora}. We also use a weight decay of $0.01$ for LoRA-based methods. We use the official code for LoRA. \footnote{\url{https://github.com/microsoft/LoRA}}

We set the infix length for all prompt tuning methods to $8$ because it generally provided the best performance in \citet{hu2021lora}. We search the prefix length of prompt tuning random, prefix tuning, and prefix-layer tuning within $\{8, 64, 256\}$. We use the same prefix length as prompt tuning random for p-tuning because both similarly operates at the embedding level. For prompt tuning, we use the same hyperparameters as tuned for prompt tuning random. 

For P-Tuning, we initialize the prompt template token embeddings with a dimension of $b$. The initialized prompt tokens are passed through a BiLSTM (separately for prefix and infix but with shared parameters) with a total hidden size of $2 \times \left \lfloor{\frac{d}{2}}\right \rfloor$ (where $d$ is the pre-trained model emebdding dimensions). We then use two affine layers with a GELU activation \cite{Hendrycks2016GaussianEL} in-between to transform the concatenation of the forward and backward hidden states (total dimension $2 \times \left \lfloor{\frac{d}{2}}\right \rfloor$) into a dimension $d$. The intermediate layer also has a dimension of $d$. We search $b$ within $\{d, \left \lfloor{\frac{d}{2}}\right \rfloor , \left \lfloor{\frac{d}{4}}\right \rfloor\}$. 

Also, during prefix tuning, we use a 2-layered MLP with GELU activation in-between to transform the prompt template embeddings from some dimension $b$ to dimension $d$. We use an intermediate layer dimension of $d$ for the MLP. Again, for prefix tuning too, we search for $b$ in $\{d, \left \lfloor{\frac{d}{2}}\right \rfloor , \left \lfloor{\frac{d}{4}}\right \rfloor\}$. We use an MLP layer because it was used by \cite{li2021prefix}. We also tuned and trained a version without the MLP layer but did not observe better performance than prompt tuning on iBLEU and SARI.

For RAPT and NC-RAPT, we set the length of $d_{1:m}$ as $x-8$ (where the value of x is the same as the total prefix length which is searched and tuned in prompt tuning and prompt tuning random). We set the length of $p_{1:s}$ (in RAPT and NC-RAPT) as $8$ (same as infix length). Thus, we keep both $p_{1:s}$ and $q_{1:t}$ small (length $8$) in RAPT and NC-RAPT whereas the majority of the parameters are concentrated on the global prefix $d_{1:m}$ ($x-8$ length) as we discussed before.

In all cases, we use AdamW \cite{loshchilov2018decoupled} as the optimizer. We also use a linear schedule with warmup for $100$ steps (we use the \verb+get_linear_schedule_with_warmup()+ function from Transformers library \cite{wolf2020transformers}), a gradient norm clipping with a maximum of $1$, a batch size of $32$, and a maximum decoding length of $n+100$ where $n$ is the size of the input prompt (which includes the prefix tokens, infix tokens, the input to be paraphrased, and all retrieved examples if any). The selected hyperparameters for each approaches from the search are provided in Table \ref{table:hyperparemeters}.   

During training, we use a maximum epoch of $30$ with early stopping. We set the early stopping patience as $3$. Model selection during training is done based on validation loss. The models are trained and tuned on single Tesla V100 32GB GPUs. Gradient accumulation is used to maintain the effective batch size as $32$.

%\subsection{Discussion}
%In Table \ref{table:rapt ablation}, we show that KNN-based example retrieval is crucial for the performance of RAPT as opposed to random retrieval. This may suggest that one weakness of the method is the RAPT-based methods are that they would not perform as well on a testing dataset where the examples are semantically distant from anything in the training dataset. 

\section{Related Work}
\textbf{Paraphrase Generation} - Traditionally rule-based systems were used for paraphrase generation \cite{mckeown1983paraphrasing, kozlowski2003generation, hassan2007unt}. \citet{quirk2004monolingual, zhao2008combining} used Statistical Machine Translation (SMT) methods for paraphrasing. More recent works typically utilize Seq2Seq models for paraphrase generation \cite{prakash2016neural, cao2017joint, zhao2018integrating, egonmwan2019transformer}. \citet{mallinson2017paraphrasing} proposed bilingual pivoting for paraphrasing. \citet{li2018paraphrase} utilized deep reinforcement learning to advance paraphrase generation whereas \citet{du2019empirical} utilized imitation learning. \citet{gupta2018deep} incorporated a variational autoencoding strategy to generate multiple diverse paraphrases. \citet{kumar2019submodular} proposed a submodular optimization-based decoding method to generate diverse paraphrases. \citet{cao-wan-2020-divgan} also improved the novelty of paraphrases through a GAN augmented with a diversity loss. \citet{Park2019paraphrase} used counterfactual debiasing to generate a diverse paraphrases. \citet{lin2021pushing} proposed multi-round paraphrase generation for improved diversity. \citet{fu2019paraphrase} proposed a latent-variable model grounded by bag-of-words from target sentences for paraphrase generation. %\citet{huang2019dictionary} proposed dictionary-guided editing for the task. %\citet{fabre2021neural} introduced a neural search-based technique for the task. \citet{zhang2021dont} used special tag tokens for selective paraphrase generation. 
\citet{witteveen-andrews-2019-paraphrasing} use GPT2 for paraphrase generation. \citet{liu-etal-2020-unsupervised, west2021reflective} proposed novel unsupervised paraphrasing strategies.

Similar to our novelty-controlled generation, there are a few approaches \cite{iyyer2018adversarial, li2019decomposable, chen2019controllable, kumar2020syntax, goyal2020neural, huang2021generating, hosking2021factorising} focusing on more controllable paraphrase generation. Although these approaches can provide extra control over different aspects (eg. granularity or syntactic templates) of paraphrase generation, they do not explicitly or directly help us in controlling novelty. Moreover, most of these approaches require specialized architectures which cannot be straightforwardly utilized in the adaptation of common pre-trained language models.%In contrast, our prompt-tuning-based approach is fairly model-agnostic. 

\setlength{\parskip}{0.3em}

\noindent \textbf{Prompt Tuning} - Initial work on prompt tuning focused on discrete selection of prompt template tokens \cite{jiang2020can, schick2021exploiting, shin2020autoprompt, schick2021small}. Some of the newer works \cite{li2021prefix, liu2021gpt, lester2021power, hu2021lora}, instead, directly tuned the prompt template token embeddings and/or intermediate layer states in the continuous space; often achieving better performance than the former strategy \cite{liu2021gpt}. Our approach follows the latter direction.  

\noindent \textbf{Retrieval Augmentation} - \citet{hashimoto2018retrieve} introduced a retrieve-and-edit framework for structured output prediction. \citet{kazemnejad2020paraphrase} built upon \citet{hashimoto2018retrieve} by using retrieved examples to augment paraphrase generation. However, their approach is not integrated with prompt tuning and uses a specialized architectural which cannot be easily utilized in adapting a generic pre-trained language model without introducing pre-training-fine-tuning discrepancies. Similar to our work, \citet{liu2021good, gao-etal-2021-making} augmented the prompts for pre-trained models using kNN-based retrieved examples. However, unlike our work, they either use manual prompts or a separate model for discrete prompt prediction (instead of tuning the prompts directly in a continuous space) while focusing on few-shot settings. Also, they did not explore paraphrase generation. To the best of our knowledge, our work is the first to integrate kNN-based example retrieval with prompt tuning, in a continuous space \cite{li2021prefix, lester2021power}, for paraphrase generation in a standard (non-few-shot) supervised training setup while, at the same time, incorporating specialized prompts for novelty controlled generation.
\setlength{\parskip}{0.0em}
\section{Conclusion}
\begin{comment}
In this paper, we propose a prompt-tuning strategy which XXX, and introduce a diversity loss to the proposed prompt-tuning strategy for the .
The experimental results demonstrate that our approach improves the performance on the task of paraphrasing on \todo{jishnu}{how many dataset?} dataset.
In particular, by incorporating the diversity, improves the diversity of paraphrasing while keeps semantic similarity 
In the further, we aim at \todo{jishnu, shuyi, yong}{add todo}
\todo{adding brainstormed future work here to discuss}
\\
Future work:
Other large language models: As future work, we are interested in applying our proposed model to other large pre-trained language models such as:... , to investigate how performance varies with different model architecture and model size. 
Auto-generated threshold for novelty controls: so far we have set the threshold for novelty controlled paraphrase generation manually. In the future we can further explore ideas to generate the thresholds automatically, such as:...
Conditional prompt tuning for other downstream tasks: We can further generalize the idea about the conditional prompt tuning  to the other downstream tasks such as:...
\end{comment}
In this paper, we propose RAPT as a parameter-efficient retrieval augmented prompt tuning setting for enhanced paraphrase generation. Building up on RAPT, we further introduce NC-RAPT for novelty-controlled paraphrase generation. Our experimental results from four datasets confirms the effectiveness of our methods. As future work, we are interested in applying our proposed approaches to other large pre-trained language models to investigate how performance varies with different model architectures and size. We would also like to explore RAPT for other downstream tasks like semantic parsing, natural language inference, named entity recognition etc. Another avenue for future research, would be testing the effectiveness of the paraphrases generated by the proposed approaches for data augmentation.

%\newpage
\section{Acknowledgments}
We would like to thank Edgar Meij, Srivas Prasad, Nimesh Ghelani and the anonymous reviewers for their constructive feedback and suggestions. We also thank Mounica Maddela for the valuable discussions. %This work was completed in partial fulfillment for the PhD degree of the first author.
%\section{Acknowledgments.}
%XXX

%\bibliographystyle{aaai22}
\bibliography{ref}

\newpage 

%\appendix
%\subfile{appendix}
\end{document}